\pdfoutput=1

\documentclass[11pt]{article}

\usepackage[]{ACL}

\usepackage{times}
\usepackage{latexsym}

\usepackage[T1]{fontenc}

\usepackage[utf8]{inputenc}

\usepackage{microtype}

\usepackage{inconsolata}

\usepackage{multirow}
\usepackage{soul} 
\usepackage{booktabs}
\usepackage{graphicx}
\usepackage{subcaption}
\usepackage{graphics}
\usepackage{xspace}
\usepackage{amsmath}
\usepackage{amssymb}  
\usepackage{adjustbox}
\usepackage{colortbl}
\usepackage{tikz}
\usepackage{pifont}

\newcommand{\cmark}{\ding{51}}%

\usepackage{todonotes}

\newcommand{\gpt}{{\textsc{GPT-3}}\xspace}
\newcommand{\park}{{\textsc{Wt-Lex}}\xspace}
\newcommand{\basic}{{\textsc{Basic}}\xspace}
\newcommand{\textbook}{{\textsc{Textbook}}\xspace}
\newcommand{\wordlist}{{\textsc{Wordlist}}\xspace}
\newcommand{\itemdesc}{{\textsc{ItemDesc}}\xspace}

\newcommand{\bothaltitems}{{\textsc{BothAltItems}}\xspace}
\newcommand{\altpos}{{\textsc{AltPos}}\xspace}
\newcommand{\altneg}{{\textsc{AltNeg}}\xspace}

\newcommand{\qcut}{\begin{small}{\textsc{Macro F1}}\end{small}\xspace}
\newcommand{\agr}{\begin{small}{\textsc{AGR}}\end{small}\xspace}
\newcommand{\ope}{\begin{small}{\textsc{OPE}}\end{small}\xspace}
\newcommand{\ext}{\begin{small}{\textsc{EXT}}\end{small}\xspace}
\newcommand{\con}{\begin{small}{\textsc{CON}}\end{small}\xspace}
\newcommand{\neu}{\begin{small}{\textsc{NEU}}\end{small}\xspace}

\newcommand\nnfootnote[1]{%
  \begin{NoHyper}
  \renewcommand\thefootnote{}\footnote{#1}%
  \addtocounter{footnote}{-1}%
  \end{NoHyper}
}

\title{Systematic Evaluation of GPT-3 for Zero-Shot Personality Estimation}

\author{Adithya V Ganesan$^{*1}$, Yash Kumar Lal$^{*1}$, August Håkan Nilsson$^2$ \\ \bf H. Andrew Schwartz$^1$ \\ $^1$Stony Brook University, NY,  $^2$Oslo Metropolitan University, Norway \\ \tt \{avirinchipur, ylal\}@cs.stonybrook.edu \\
} 

\begin{document}
\maketitle
\begin{abstract}
Very large language models (LLMs) perform extremely well on a spectrum of NLP tasks in a zero-shot setting.
However, little is known about their performance on human-level NLP problems which rely on understanding psychological concepts, such as assessing personality traits. 
In this work, we investigate the zero-shot ability of \gpt to estimate the Big 5 personality traits from users' social media posts.
Through a set of systematic experiments, we find that zero-shot \gpt performance is somewhat close to an existing pre-trained SotA for broad classification upon injecting knowledge about the trait in the prompts. 
However, when prompted to provide fine-grained classification, its performance drops to close to a simple most frequent class (MFC) baseline.
We further analyze where \gpt performs better, as well as worse, than a pretrained lexical model, illustrating systematic errors that suggest ways to improve LLMs on human-level NLP tasks.
\end{abstract}



\section{Introduction}
Human-level NLP tasks, rooted in computational social science, focus on the link between social or psychological characteristics and language. 
Example tasks include personality assessment~\cite{mairesse-walker-2006-automatic,Kulkarni2017LatentHT,lynn-etal-2020-hierarchical}, demographic estimation~\cite{sap-etal-2014-developing, PreotiucPietro2018UserLevelRA}, and mental health-related tasks~\cite{coppersmith-etal-2014-quantifying, Guntuku2017DetectingDA, matero-etal-2019-suicide}. 
Although using LMs as embeddings or fine-tuning them for human-level NLP tasks is becoming popular~\cite{v-ganesan-etal-2021-empirical,butala-etal-2021-team,yang-etal-2021-learning-answer}, very little is known about zero-shot performance of LLMs on such tasks.    
\nnfootnote{*These authors contributed equally}


In this paper, we test the zero-shot performance of a popular LLM, \gpt, to perform personality trait estimation.
We focus on personality traits because they are considered the fundamental characteristics that distinguish people, persisting across cultures, demographics, and time~\cite{costa1992normal, costa1996mood}. 
These characteristics are useful for a wide range of social, economic, and clinical applications such as understanding psychological disorders~\cite{khan2005personality}, choosing content for learning styles~\cite{komarraju2011big} or occupations~\cite{kern2019social},  and delivering personalized treatments for mental health issues~\cite{bagby2016role}.
Focusing on zero-shot evaluation of \gpt on these fundamental characteristics forms a strong benchmark for understanding how much and what dimensions of traits \gpt encodes out-of-the-box.
Further, while fine-tuned LMs have only had mixed success beyond lexical approaches~\cite{lynn-etal-2020-hierarchical, kerz-etal-2022-pushing}, using zero-shot capable LLMs could help lead to better estimates. 
 



The NLP community has a growing interest in understanding the capabilities and failure modes of LLMs~\cite{wei2022emergent,Yang2021AnES}, and we explore questions that surround LLMs in the context of fundamental human traits of personality. 
Zero-shot performance can depend heavily on the explicit information infused in the prompt~\cite{lal-etal-2022-using}.
Personality, defined by information in its well-established questionnaire tests, presents new opportunities for information infusion. 

Our \textbf{contributions} address: \textbf{(1)} what information about personality is useful for \gpt, \textbf{(2)} how its performance compares to current SotA, \textbf{(3)} the relation between ordinality of outcome labels with performance and \textbf{(4)} whether \gpt predictions stay consistent given similar external knowledge.

\section{Background}

Psychological traits are stable individual characteristics associated with behaviors, attitudes, feelings, and habits \cite{apaDicitionary23}. 
The ``Big 5'' is a popular personality model that breaks characteristics into five fundamental dimensions, validated across hundreds of studies across cultures, demographics, and time~\cite{costa1992normal, mccrae1992introduction}.
The approach is rooted in the \textit{lexical hypothesis} that the most important traits must be encoded in language~\cite{goldberg1990alternative}.
We investigate all five factors from this model:
openness to experience (\ope -- intellectual, imaginative and open-minded), conscientiousness (\con -- careful, thorough and organized), extraversion (\ext -- energized by social and interpersonal interactions), agreeableness (\agr -- friendly, good natured, conflict avoidant) and neuroticism (\neu -- less secure, anxious, and depressive).

LLMs like PaLM \cite{chowdhery2022palm} have shown significant improvement in performance on various NLP tasks~\cite{wei2022chain, Suzgun2022ChallengingBT}, even without finetuning. 
There is a growing body of work investigating one of the ubiquitous LLMs, \gpt, under different settings~\cite{wei2022emergent, shi2022language, bommarito2023gpt}. 
Inspired by this, we systematically study the ability of \gpt to perform personality assessment under zero-shot setting.
Following evidence that incorporating knowledge about the task can improve performance~\cite{vu-etal-2020-predicting, yang-etal-2021-learning-answer, lal-etal-2022-using}, we evaluate the impact of three different types of knowledge to determine which type improves personality estimation.


Modeling personality traits through natural language has been extensively studied using a wide range of approaches, from simple count-based models~\cite{pennebaker2003words, Golbeck2011PredictingPW} to complex hierarchical neural networks~\cite{read2010neural, yang-etal-2021}. 
Finetuning LMs has become the mainstream approach for this task only recently~\cite{v-ganesan-etal-2021-empirical}.
With the advent of \gpt, zero- or few-shot settings have become the primary approach to leverage LLMs in other NLP applications, but are yet untested for personality estimation.

\begin{table*}[!tbh]
\centering
\begin{tabular}{|l|ccccc|c|}
\hline
Model  & \ope & \con & \ext & \agr & \neu & Avg\\
\hline
\multicolumn{7}{|c|}{Benchmarks}\\
\hline
MFC & 0.352 & 0.427 & 0.411 & 0.372 & 0.333 & 0.379 \\
\park (\citeauthor{park-etal-2015}) & \textbf{0.492} & 0.393 & 0.516 & \textbf{0.609} & \textbf{0.578} & \textbf{0.518} \\\hline
\multicolumn{7}{|c|}{Zero-Shot \gpt}\\
\hline
\basic & 0.329$^\dag$ & 0.385 & 0.521 & 0.435$^\ddag$ & 0.333$^\ddag$ & 0.400 \\
\quad\textbook & 0.328$^\dag$ & 0.401 & 0.496 & 0.506$^*$ & 0.364$^\ddag$ & 0.419 \\
\quad\wordlist & 0.366$^\dag$ & 0.457 & 0.445 & 0.544 & 0.393$^\ddag$ & 0.441 \\
\quad\itemdesc & 0.342$^\dag$ & \textbf{0.521}$^\dag$ & \textbf{0.569} & 0.488$^\dag$ & 0.349$^\ddag$ & 0.454 \\
\hline

\end{tabular}
\caption{\qcut scores for different kinds of knowledge added to the prompt. \textbook refers to adding the definition of the trait as described in~\citet{roccas-etal-2002}, \wordlist refers to adding the top 5 positively and negatively correlated unigrams with the trait reported by~\citet{schwartz2013personality}, \itemdesc refers to adding the items that were a part of the personality questionnaire (\autoref{tab:survey_items}). \park refers to the SotA model described in \S \ref{sec:expt}. The findings indicate a statistically significant distinction when compared to the \park model, with significance levels of $p<0.05$ (*), $p<0.01$ ($\dag$), and $p<0.001$ ($\ddag$).}
\label{tab:main_results}
\end{table*}


\section{Dataset}
\label{sec:data}
To get a sample of language associated with personality, we followed the paradigm set forth in~\citet{jose2022using} whereby consenting participants shared their own Facebook posts along with taking a battery of psychological assessments, including the big five personality test~\citep{donnellan2006mini, Kosinski2013PrivateTA}. 
The dataset comprises of 202 participants with outcomes of interests who had also shared their Facebook posts.
First, we filter the data to only include user posts from the last year of data collection ~\cite{eichstaedt2018facebook}.
Next, we only retain users for whom we have exactly 20 Facebook posts, similar to the approach described in other human-level NLP works~\cite{lynn-etal-2020-hierarchical, matero-etal-2021-melt-message}.
Finally, we anonymize the data by replacing personable identifiable information using SciPy's ~\cite{2020SciPy-NMeth} NER model.
We also remove phone numbers and email IDs using regular expressions.
Finally, we are left with anonymized Facebook posts for 142 users and their associated 5 personality traits. This population (all from US) has a gender ratio of 79:18:3 (female:male:others). The age ranged from 21 to 66 (median=37). 
The big 5 personality trait scores fall in the continuous range of [1, 5]. 
We discretize the outcome values into the desired number of bins/classes using a quantile discretizer (in Pandas). We explain why we choose to discretize the outcome values in \S \ref{sec:expt}.

\section{Experimental Design}
\label{sec:expt}


In this work, \gpt is evaluated in a zero-shot setting. 
We frame the problem of personality prediction as classifying the degree (i.e. high/low or high/medium/low) to which a person exhibits a trait. 
Ideally, because the big 5 are considered continuously valued variables~\citep{mccrae1989reinterpreting}, one would model as a regression task, but we found this simplification to classification necessary to get any meaningful insights from \gpt's zero-shot capability. 
We also investigate the degradation of performance for tertiary classification instead of binary in \S \ref{sec:results}. 



We devise a simple, reasonable prompt (\basic)\footnote{Examples of all prompts are in Appendix \autoref{fig:ext_knowl}.} to first estimate the ability of \gpt to predict the Big 5 personality traits.
Building on this, we investigate whether adding external knowledge about these traits helps the model perform better.
We use three types of knowledge: \textbf{(1) \textbook}: a concise definition of these traits from~\citet{roccas-etal-2002}, \textbf{(2) \wordlist}: frequent and infrequent words\footnote{We use the wordlist from~\citet{schwartz2013personality}.} used by people exhibiting those traits, and \textbf{(3) \itemdesc}: survey items\footnote{See Appendix \autoref{tab:survey_items} for detailed item descriptions.} (a positive and a negative) users responded to, based on which their personality scores were estimated.

\paragraph*{Baseline and Evaluation.} 

The baseline, \park, is a ridge regression model from~\citealt{park-etal-2015} trained on dimensionally reduced feature set of n-grams and LDA-based topics extracted from~\citet{Kosinski2013PrivateTA} Facebook data. The number of parameters in this model is orders of magnitude less than \gpt.
Even complex neural models~\cite{lynn-etal-2020-hierarchical} have been unsuccessful to surpass its performance. 
\park also produces predictions in the continuous scale within the range of [1, 5]. 
In order to make a fair comparison with \gpt, we perform the quantile discretization described in \S \ref{sec:data} and calculate \qcut. 
We evaluate the predictions using macro F1 scores.

\section{Results} 
\label{sec:results}

\autoref{tab:main_results} shows \gpt's performance on different personality traits, with and without knowledge.
We find that \itemdesc prompts the best performance with \gpt on average.
Surprisingly, the model is able to directly use survey items (\itemdesc) to predict \ext and \con the best.
Utilizing these is hard since it requires relating abstract concepts described in these survey items to the ecological language in the posts.
The top frequent and infrequent words (\wordlist) help model perform the most on \agr, \ope and \neu.
We hypothesize that simple, lexical cues are more helpful here since it is easier to draw relations from the surface form in posts.
We also note that estimating \neu is difficult for the model, which also is difficult for humans to estimate in zero-acquaintance contexts, ~\cite{kenny1994interpersonal}, including estimating neuroticism from Facebook profiles. 
Overall, \gpt's predictions are heavily biased towards predicting individuals to be high openness and low in neuroticism.

We also tried incorporating all types of knowledge into a prompt and found that performance dropped below \basic. 
However, combining knowledge types involves non-trivial decisions such as the order of knowledge types and its composition. 
We leave this to future work.



Using \itemdesc, we establish the best possible \gpt performance for personality estimation. 
Although \gpt's average performance over all traits is still lower than \park, it outperforms the MFC baseline. 
Prior work~\cite{v-ganesan-etal-2022-wwbp, matero-etal-2022-understanding} has shown dimensions of mental health constructs and personality traits being captured through language use patterns in LMs. 
\gpt's performance in zero-shot setting provides reasonable evidence to believe that language patterns associated with these traits are encoded in its embedding space as well.

\section{Analysis}
\label{sec:analysis}
To better understand the utility of \gpt for personality estimation, we analyze the effect of \textbf{(1)} problem framing, and \textbf{(2)} effect of survey items.
Furthermore, we perform error analysis of \gpt to suggest avenues for improvement.


\paragraph*{Problem Framing.} When personality estimation is framed as a binary classification, \gpt is worse than SoTA on average in a zero-shot setting. 
Upon looking closer, we note that it is the best model for 2 out of the 5 traits. 
However, these observations are made in a simplified two-class setting, whereas the big 5 personality model produces a real valued outcome.
In order to assess \gpt's practical viability, we prompt it (\itemdesc) to provide more fine-grained predictions by presenting trait estimation as a three-class classification problem. 

\begin{table}[!th]
\centering
\adjustbox{max width=\columnwidth}{%
\begin{tabular}{|c|c|c|c|c|c|c|}
\hline
  \# class  & \ope & \con & \ext & \agr & \neu & Avg\\\hline
2 & 0.342 & 0.521 & 0.569 & 0.488 & 0.349 &   
0.454 \\
3 & 0.141 & 0.288 & 0.240 & 0.160 & 0.320 & 0.230 \\\hline
\end{tabular}}
\caption{\qcut scores of classifying the outcomes into varying number of classes using \gpt. We find a sharp drop in performance on increasing the number of classes from 2 to 3. Hence, framing personality estimation as a binary classification is the simplest for \gpt}
\label{tab:framing}
\end{table}



\autoref{tab:framing} shows that problem framing has a major impact on \gpt performance for all traits.
Three class framing of the problem is harder than the binary framing which is evident from \gpt's drop in performance (0.229) to close to MFC (0.212).
This trend indicates that \gpt is ineffective in performing more fine-grained prediction tasks and consequently regression, which is the natural way to estimate the Big 5 traits. 
Clearly, \gpt is yet unsuited for fine-grained personality estimation.

\paragraph*{Consistency with Survey Items.} 
The standard questionnaire used to create the dataset had a total of 4 survey items per trait (2 positive and 2 negative).
For \itemdesc, we use one positive and one negative item to describe each trait (see \autoref{fig:ext_knowl}).
To investigate whether \gpt performance can be attributed to specific items in the prompt, we perform \itemdesc with all possible combinations of a positive and a negative survey item for all traits.

\begin{table}[!h]
\small
\centering
\begin{tabular}{|c|c|}
\hline
  & Avg\\\hline
\itemdesc & 0.454 \\
\bothaltitems & 0.448 \\
\altpos & 0.430 \\
\altneg & 0.448 \\
\hline
\end{tabular}
\caption{\qcut scores for different pairs of positive and negative survey items combinations. \autoref{tab:survey_items} in Appendix contains the survey items that correspond to these four combination labels. 
}
\label{tab:consistency}
\end{table}

\autoref{tab:consistency} shows that there is no meaningful difference in performance when provided different item combinations. 
This shows that \gpt is not sensitive to the items of the personality questionnaire. 
This is in line with data in ~\autoref{tab:factors}, which shows that factor loading values \cite{fabrigar2011exploratory} of these item combinations have similar powers to distinguish the corresponding traits.





\begin{figure}[!t]
    \centering
    \begin{minipage}[b]{0.235\textwidth}
        \centering
        \includegraphics[width=\textwidth]{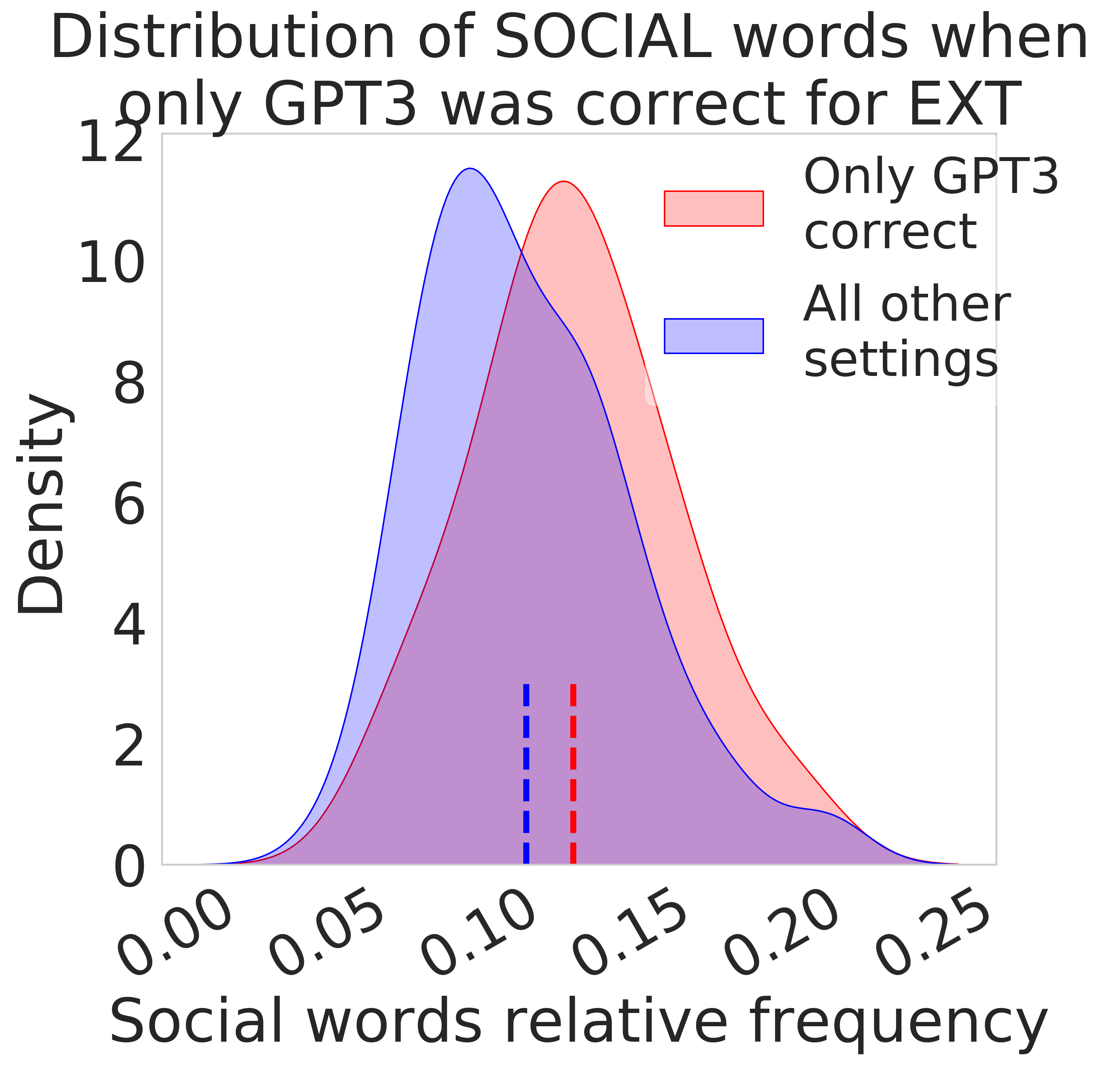}
        \subcaption{}
        \label{fig:err_analysis_a}
    \end{minipage}
    \begin{minipage}[b]{0.235\textwidth}
        \centering
        \includegraphics[width=\textwidth]{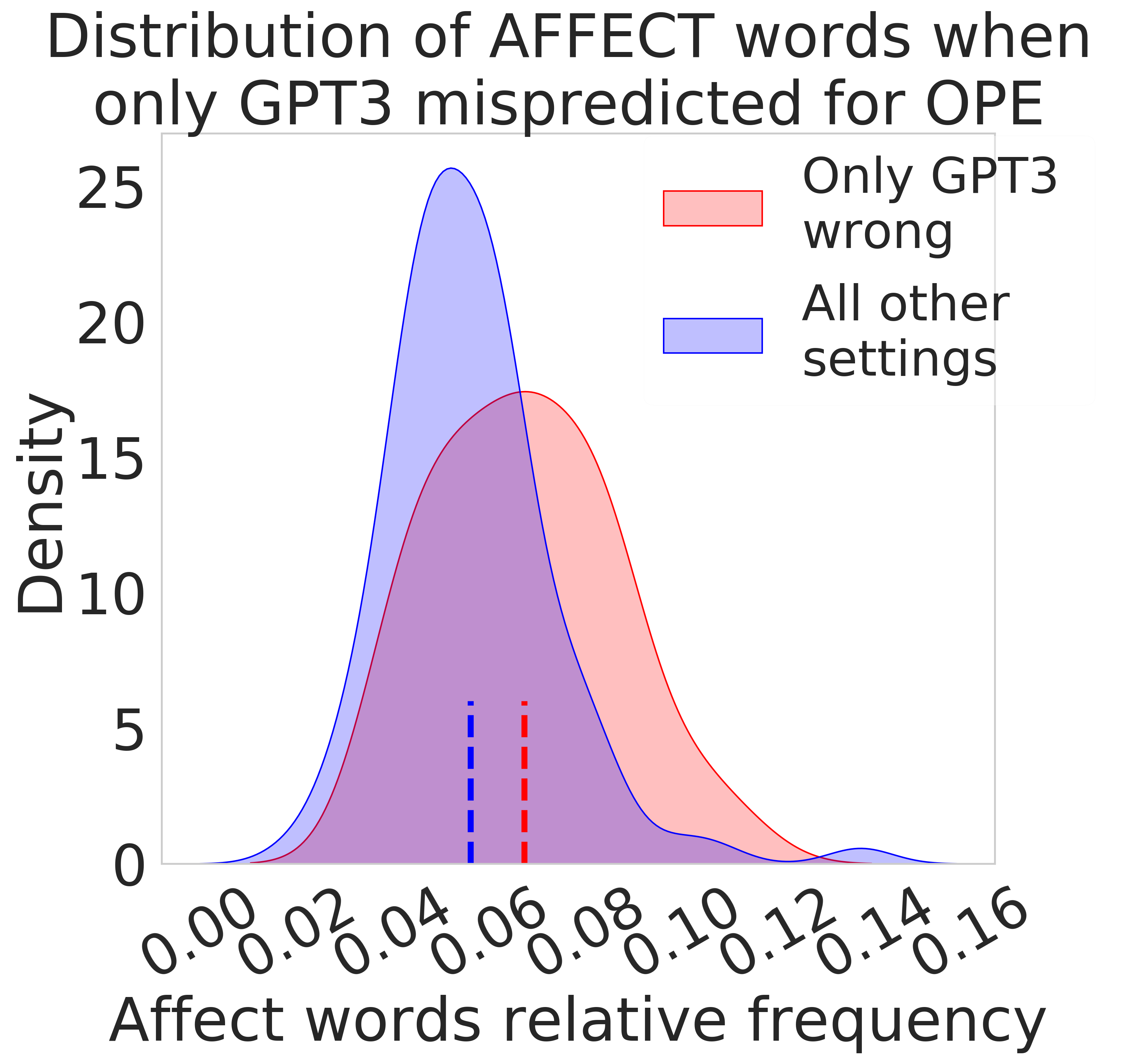}
        \subcaption{}
        \label{fig:err_analysis_b}
    \end{minipage}
    \caption{(a) SOCIAL words distributions compared for \gpt and \park under two prediction settings: (1) only \gpt correct, and \park incorrect, and (2) both models correct or \gpt incorrect. (right) AFFECT words distributions compared for \gpt and \park under two prediction settings: (1) only \gpt incorrect, and \park correct, and (2) both models incorrect or \gpt correct.}
    \label{fig:err_analysis}
\end{figure}

\paragraph*{Error Analysis.} Finally, we examine the linguistic variables that account for the errors in \gpt and the areas where it excels as compared to a traditional, lexical-based technique \park. 
\autoref{fig:err_analysis_a} shows the distributions of SOCIAL words ~\citep{Tausczik2010ThePM} between users that were correctly predicted by only \gpt and the users that were either misclassfied by \gpt or correctly predicted by \park for \ext task. 
SOCIAL words are better captured by LLMs probably owing to its ability to produce contextualized embeddings.
\autoref{fig:err_analysis_b} depicts the distributions of AFFECT words between the users that were misclassified only by \gpt and the users that were either correctly classified by \gpt or \park misclassifies for \ope task\footnote{We also looked at the differences in other LIWC categories for \ext and \ope tasks measured using Cohen's d~\cite{cohens} and logs odds ratio with informative dirichlet prior~\cite{monroe2008fightin} that offers more explanations for the errors and correctness of \gpt in \autoref{appendix_err_analysis}.}.

\section{Conclusion}

We performed a systematic investigation of \gpt's zero-shot performance on personality estimation.
While using a simple prompt did not yield strong performance, injecting knowledge about the traits themselves led to significant improvement.
Even so, it falls short of using a strong, extensively-trained, supervised model (\park). 
Further, we find that it is much harder for \gpt to provide more fine-grained predictions (when asked to select between 3 labels instead of 2), suggesting that LLMs may not be as capable at making dimensional estimates about personality.
Our systematic investigation helps understand \gpt's zero-shot capabilities for a human-level NLP task, contextualizing its failure modes and showing avenues for LLM improvements.

\section*{Ethics Statement}

Our work seeks to advance interdisciplinary NLP-psychology research for understanding human attributes associated with language.
This research is intended to inform Computational Social Science researchers about the ability of LLMs to estimate psychological rating scales as well as for LLM researchers to understand types of psychological information that LMs capture. 
We intend for our work on personality trait assessments to have an impact on social, NLP, and clinical use cases to improve the well-being of people. 
We strongly condemn malevolent adoption of these technologies for targeted advertising, directed misinformation campaigns, and other malicious acts that could have potential harms on mental health. 

If used for clinical practice, we strongly recommend that any use of LLM-based personality estimates be overseen by clinical psychology experts. 
During trials, models should be extensively tested for their failure mode rates (e.g. False-positive vs False-negative rates), and error disparities~\cite{shah-etal-2020-predictive}. 

This interdisciplinary computer science, psychology, and health study had extensive privacy \& ethical human subjects research protocols. 
All procedures were approved by an academic institutional review board. 
All contributors are certified to perform human subject research, and took steps and precautions while collecting and analyzing data to keep participants protected. 
The Facebook posts shared by consenting users were anonymized as described in \S \ref{sec:data} to prevent the participants from being identified. 

\section*{Limitations}
The Big 5 personality trait model measures the fundamental dimensions of human on a continuous scale. 
This real valued representation preserves more information and is more descriptive of inter-individual differences. 
While we acknowledge that the binary classification of Big 5 traits fails the purpose of the model, it is a necessary simplification to understand the ability of LLMs to perform personality assessment. 
Our investigation shows potential to improve the practical utility of LLMs in personality estimation.

Despite the strong results from existing works in support of in-context learning and larger message history for better performance, we were limited by the significant multiplicative cost these experiments entailed, as the \gpt API is billed based on token usage. 
Further, since each user's post history is typically long, it is infeasible to experiment with all in-context learning options due to \gpt's context window size limitation.
This is worthy of exploration, to understand the sample efficiency of \gpt and the impact of post history on its performance.

\section*{Acknowledgement}

This work wouldn't have been possible without the support of AVG's and YKL's dear friends, Swanie Juhng, Aakanksha Rajiv Kapoor, Aravind Parthasarathy, Aditya Krishna, Akshay Bharadhwaj, and Somadutta Bhatta, who provided OpenAI API keys for running the experiments. We would also like to extend our gratitude to Matthew Matero, Niranjan Balasubramanian, Harsh Trivedi and Sid Mangalik for providing valuable feedback. AVG, AHN, and HAS were supported in part by NIH grant R01-AA028032 and YKL was supported by DARPA. 

\bibliography{acl2023}
\bibliographystyle{acl_natbib}

\clearpage

\appendix

\section{GPT-3}

\subsection{GPT-3 settings}

We used a temperature of 0.0 for all the experiments to select the most likely token at each step,
as this setting allows for reproducibility.

{\small
\begin{verbatim}
response = openai.Completion.create(
            model="text-davinci-003",
            prompt=prompt,
            temperature=0,
            max_tokens=1,
            top_p=1.0,
            frequency_penalty=0.1,
            presence_penalty=0.0
)
\end{verbatim}}

We restricted the model outputs to just one token.
Only ``Yes" or ``No" are considered valid answers for our binary classification task.
For the 3-class classification, ``High", ``Medium" and ``Low" are considered valid answers.

For one data point in the \wordlist \ext experiment, the model output was a newline character instead of Yes/No. By adding another newline to the prompt, we were able to get it to generate an answer (in this case, No).
For one data point in the \basic \ope experiment, the model output contained irrelevant tokens instead of High/Medium/Low. By adding another 2 newlines to the prompt, we were able to get it to generate an answer (in this case, High).

\subsection{Prompt Design}

For our binary classification task, we used the following prompt template:
\begin{verbatim}
Read the stream of Facebook posts from a 
user below. Each newline represents a new 
post. The posts are in order of date, the 
last one is the most recent.
{messages}
{knowledge} Given these messages from a 
user, is this user {trait} according to 
the Big 5 personality traits? Select 
between yes or no
\end{verbatim}

A user's posts are concatenated with the most recent post presented at the end to fill the \textit{messages} field.
Options for \textit{trait} are agreeable, extraverted, open to experiences, neurotic, and conscientious.

For our 3-class problem framing, we used the following prompt template:
\begin{verbatim}
Read the stream of Facebook posts from a 
user below. Each newline represents a new 
post. The posts are in order of date, the 
last one is the most recent.
{messages}
{knowledge} Given these messages from a 
user, rate their {trait}. The options on 
the scale are low, medium, high.
{trait}:
\end{verbatim}

\begin{figure*}[!tbh]
    \centering
    \includegraphics[width=\textwidth]{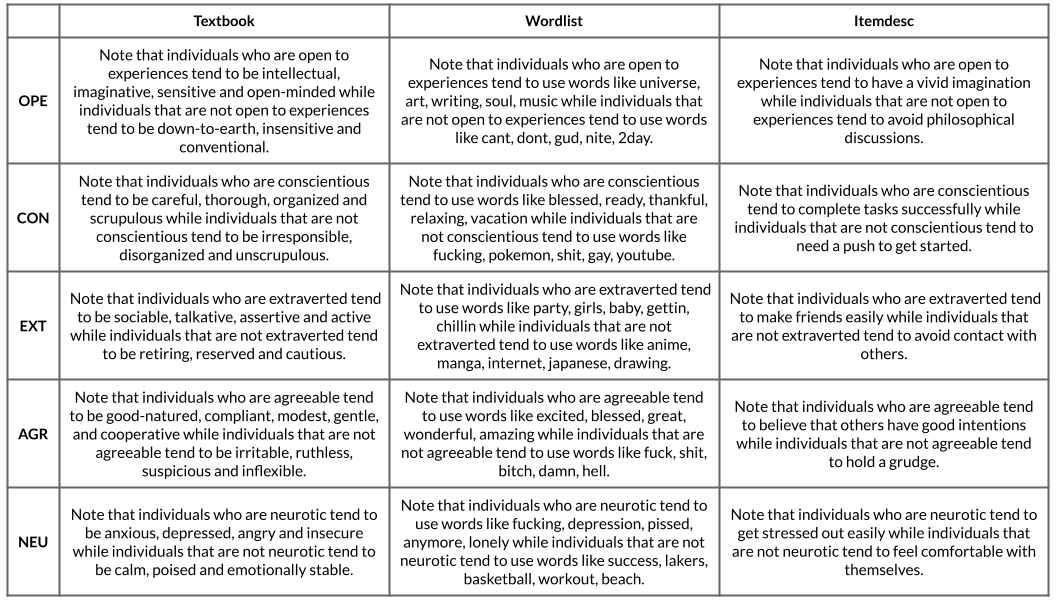}
    \caption{Different types of knowledge used for each trait in the prompt.}
    \label{fig:ext_knowl}
\end{figure*}

Options for \textit{trait} are agreeableness, extraversion, openness to experiences, neuroticism, and conscientiousness.
The different types of knowledge injected into the prompt for each personlity trait can be found in \autoref{fig:ext_knowl}.

\section{Glossary}

We include the survey items from the questionnaires used in the study to collect data from consenting users along with their associated personality trait in \autoref{tab:survey_items}, as well as the categories of language from the LIWC error analysis model in \autoref{tab:liwc}.

\begin{table}[!tbh]
    \centering
    \small
    \adjustbox{max width=\columnwidth}{%
    \begin{tabular}{|l|l|l|}
    \hline
    Category Abbrev & Category & Examples \\
    \hline
    NUMBER & Numbers & second, thousand \\
    SOCIAL & Social Processes & mate, talk, they \\
    AFFILIATION & Affiliation & ally, friend, social\\
    YOU & 2nd Person & you, your, thou \\
    TIME & Time & end, until, season \\
    FAMILY & Family & daughter, dad, aunt\\
    PPRON & Personal Pronoun & I, them, her \\
    POSEMO & Positive Emotion & love, nice, sweet \\
    AFFECT & Affective Processes & happy, cried \\
    FRIEND & Friends & neighbor, buddy \\
    THEY & 3rd Person plural & they, their, they'd \\
    FOCUSPAST & Past Focus & ago, did, talked \\
    ACHIEVE & Achievement & success, win, better \\
    SHEHE & 3rd person singular & she, him, her \\
    NEGATE & Negation & not, never, no \\
    PRONOUN & Total Pronouns & I, them, itself \\
    \hline
    \end{tabular}}
    \caption{LIWC glossary to map the category abbreviation with its full form and a few examples for each row.}
    \label{tab:liwc}
\end{table}

\section{Error Analysis}
\label{appendix_err_analysis}

We examine where \gpt differs from \park: \textbf{(1)} performing better on \ext in  \autoref{tab:diff_analysis_ext}, and \textbf{(2)} predicting \ope worse in \autoref{tab:diff_analysis_ope}.
Results from \autoref{tab:diff_analysis_ext} suggest that \gpt encodes language categories\footnote{See \autoref{tab:liwc} for details on LIWC categories} \cite{Tausczik2010ThePM} highly predictive of \ext such as social processes (SOCIAL), group identification (AFFILIATION), and use of second person pronoun (YOU), all of which have been shown to have strong significant association with this trait \cite{schwartz2013personality}.
\gpt can disambiguate common social lexicons occurring in different contexts \cite{burdick-etal-2022-using} (e.g., "party" in the context of gathering vs political ideology), which count-based lexical models can't do.

\autoref{tab:diff_analysis_ope} indicates that \gpt fails for \ope on language reflective of social processes (SOCIAL) and affect (AFFECT).
Previous work on lexical correlates of personality showed that these categories are discussed more for users low in openness~\citep{yarkoni2010personality}, 
suggesting (together with our result) that \gpt misses the connection between these categories of language and personality. 
These are areas to improve the human-level capabilities of \gpt.

\begin{table}[!t]
    \centering
    \begin{tabular}{l|c|c}
    Category &  $d$ & $OR_{IDP}$ \\\hline
    NUMBER          & 0.699     & 0.140          \\
    SOCIAL          & 0.595     & 0.191          \\
    AFFILIATION     & 0.459     & 0.140          \\
    YOU             & 0.451     & 0.132          \\
    TIME            & 0.448     & 0.115          \\
    FAMILY          & 0.395     & 0.108          \\
    PPRON           & 0.359     & 0.104          \\
    POSEMO          & 0.341     & 0.102          \\
    AFFECT          & 0.242     & 0.061         \\
    FRIEND          & 0.217     & 0.057
    \end{tabular}
   \caption{Lexical categories that are more prevalent when \gpt performs better than \park that explain their \ext predictions. $d$: Cohen's $d$ -- standardized difference in means~\cite{cohens}; $OR_{IDP}$: log \textbf{o}dds \textbf{r}atio with informative dirichlet prior~\cite{monroe2008fightin}. }
   \label{tab:diff_analysis_ext}
\end{table}

\begin{table}[!t]
    \centering                      
    \begin{tabular}{l|c|c}
    Category & $d$ & $OR_{IDP}$ \\\hline
    THEY            & 0.701     & 0.126          \\
    FOCUSPAST       & 0.692     & 0.166          \\
    AFFECT          & 0.676     & 0.132          \\
    ACHIEVE         & 0.629     & 0.104          \\
    SOCIAL          & 0.608     & 0.168          \\
    SHEHE           & 0.588     & 0.172          \\
    PPRON           & 0.559     & 0.139          \\
    NEGATE          & 0.517     & 0.082          \\
    PRONOUN         & 0.510     & 0.105         \\
    POSEMO          & 0.482     & 0.118         
    \end{tabular}
    \caption{Lexical categories that are more prevalent when \gpt performs worse for the \ope task than \park. $d$: Cohen's $d$ -- standardized difference in means of errors~\cite{cohens}; $OR_{IDP}$: log \textbf{o}dds \textbf{r}atio with informative dirichlet prior~\cite{monroe2008fightin} on errors.}
    \label{tab:diff_analysis_ope}
\end{table}

\begin{table*}
\small
\centering
\begin{tabular}{|l|l|c|c|c|c|c|}
\hline
Trait                & Survey Item                               & Polarity     & \itemdesc  & \altpos & \altneg & \bothaltitems \\\hline
\multirow{4}{*}{OPE} & Have a vivid imagination                  & +        & \cmark &  & \cmark &  \\
                     & Avoid philosophical discussions           & -        &  & \cmark & \cmark & \\
                     & Enjoy wild flights of fantasy             & +        &  & \cmark &  & \cmark\\
                     & Do not like poetry                        & -        & \cmark &  &  & \cmark\\\hline
\multirow{4}{*}{CON} & Complete tasks successfully               & +        & \cmark & \cmark &  & \\
                     & Need a push to get started                & -        & \cmark &  & \cmark & \\
                     & Am always prepared                        & +        &  &  & \cmark & \cmark\\
                     & Shirk my duties                           & -        &  & \cmark &  & \cmark\\\hline
\multirow{4}{*}{EXT} & Do not mind being the centre of attention & +        &  &  & \cmark & \cmark\\
                     & Make friends easily                       & +        & \cmark & \cmark &  & \\
                     & Keep in the background                    & -        &  & \cmark & \cmark & \\
                     & Avoid contact with others                 & -        & \cmark &  &  & \cmark\\\hline
\multirow{4}{*}{AGR} & Hold a grudge                             & -        & \cmark & \cmark &  & \\
                     & Believe that others have good intentions  & +        & \cmark &  & \cmark & \\
                     & Cut others to pieces                      & -        &  &  & \cmark & \cmark\\
                     & Am easy to satisfy                        & +        &  & \cmark &  & \cmark\\\hline
\multirow{4}{*}{NEU} & Feel comfortable with myself              & -        & \cmark & \cmark &  & \\
                     & Often feel blue                           & +        &  & \cmark &  & \cmark\\
                     & Get stressed out easily                   & +        & \cmark &  & \cmark & \\
                     & Am not easily bothered by things          & -        &  &  & \cmark & \cmark\\\hline
\end{tabular}
\caption{Survey items from the questionnaires answered by people for Big 5 personality assessment along with the combination labels these items were a part of (referenced in \autoref{tab:consistency}.}
\label{tab:survey_items}
\end{table*}

\begin{table*}[!tbh]
\centering
\small
\begin{tabular}{|c|c|l|l|c|c|}
\hline
\textbf{Trait}        & \textbf{\begin{tabular}[c]{@{}c@{}}Item \\ Combination\end{tabular}} & \multicolumn{1}{c|}{\textbf{Positive Item}} & \multicolumn{1}{c|}{\textbf{Negative Item}} & \textbf{\begin{tabular}[c]{@{}c@{}}Factor \\ Loading\end{tabular}} & \textbf{Macro F1}             \\
\hline
                      & ~\itemdesc                                                                 & Have a vivid imagination                                                             & Do not like poetry                         & \cellcolor[HTML]{FCE5CD}0.703                                      & \cellcolor[HTML]{FCE5CD}0.335 \\
                      & ~\altneg                                                               & Have a vivid imagination                                                             & Avoid philosophical discussions            & \cellcolor[HTML]{FADBBA}0.714                                      & \cellcolor[HTML]{F8D6B1}0.342 \\
                      & ~\altpos                                                               & Enjoy wild flights of fantasy                                                        & Avoid philosphical discussions             & \cellcolor[HTML]{F8D4AF}0.720                                      & \cellcolor[HTML]{F8D6B1}0.342 \\
\multirow{-4}{*}{\ope} & ~\bothaltitems                                                              & Enjoy wild flights of fantasy                                                        & Do not like poetry                         & \cellcolor[HTML]{E69138}0.787                                      & \cellcolor[HTML]{E69138}0.374 \\
\hline
                      & ~\itemdesc                                                                 & Complete tasks successfully                                                          & Need a push to get started                 & \cellcolor[HTML]{CFE2F3}0.781                                      & \cellcolor[HTML]{3D85C6}0.521 \\
                      & ~\altneg                                                               & Am always prepared                                                                   & Need a push to get started                 & \cellcolor[HTML]{9FC4E5}0.800                                      & \cellcolor[HTML]{CFE2F3}0.457 \\
                      & ~\altpos                                                               & Complete tasks successfully                                                          & Shirk my duties                            & \cellcolor[HTML]{68A0D3}0.821                                      & \cellcolor[HTML]{A4C7E6}0.476 \\
\multirow{-4}{*}{\con} & ~\bothaltitems                                                              & Am always prepared                                                                   & Shirk my duties                            & \cellcolor[HTML]{3D85C6}0.837                                      & \cellcolor[HTML]{99C0E3}0.481 \\
\hline
                      & ~\itemdesc                                                                 &  Make friends easily                                                                  & Avoid contact with others                  & \cellcolor[HTML]{FCE5CD}0.766                                      & \cellcolor[HTML]{E69138}0.569 \\
                      & ~\altneg                                                               & \begin{tabular}[c]{@{}l@{}}Do not mind being the \\ \quad centre of attention\end{tabular} & Keep in the background                     & \cellcolor[HTML]{EAA053}0.843                                      & \cellcolor[HTML]{FADBBB}0.528 \\
                      & ~\altpos                                                               & Make friends easily                                                                  & Keep in the background                     & \cellcolor[HTML]{EA9E4E}0.846                                      & \cellcolor[HTML]{EFB374}0.551 \\
\multirow{-4}{*}{\ext} & ~\bothaltitems                                                              & \begin{tabular}[c]{@{}l@{}}Do not mind being the \\ \quad centre of attention\end{tabular} & Avoid contact with others                  & \cellcolor[HTML]{E69138}0.860                                      & \cellcolor[HTML]{FCE5CD}0.523 \\
\hline
                      & ~\altpos                                                               & Am easy to satisfy                                                                   & Hold a grudge                              & \cellcolor[HTML]{CFE2F3}0.725                                      & \cellcolor[HTML]{97BEE2}0.501 \\
                      & ~\itemdesc                                                                 & \begin{tabular}[c]{@{}l@{}}Believe that others have \\ \quad good intentions\end{tabular}  & Hold a grudge                              & \cellcolor[HTML]{B5D2EB}0.741                                      & \cellcolor[HTML]{CFE2F3}0.488 \\
                      & ~\altneg                                                               & \begin{tabular}[c]{@{}l@{}}Believe that others have \\ \quad good intentions\end{tabular}  & Cut others to pieces                       & \cellcolor[HTML]{4389C8}0.809                                      & \cellcolor[HTML]{78ABD9}0.509 \\
\multirow{-4}{*}{\agr} & ~\bothaltitems                                                              & Am easy to satisfy                                                                   & Cut others to pieces                       & \cellcolor[HTML]{3D85C6}0.813                                      & \cellcolor[HTML]{3D85C6}0.523 \\
\hline
                      & ~\altpos                                                               & Often feel blue                                                                      & Feel comfortable with myself               & \cellcolor[HTML]{FCE5CD}0.697                                      & \cellcolor[HTML]{FCE5CD}0.333 \\
                      & ~\altneg                                                               & Get stressed out easily                                                              & Am not easily bothered by things           & \cellcolor[HTML]{EBA55A}0.804                                      & \cellcolor[HTML]{E69138}0.364 \\
                      & ~\itemdesc                                                                 & Get stressed out easily                                                              & Feel comfortable with myself               & \cellcolor[HTML]{E7953F}0.829                                      & \cellcolor[HTML]{F1BB82}0.349 \\
\multirow{-4}{*}{\neu} & ~\bothaltitems                                                              & Often feel blue                                                                      & Am not easily bothered by things           & \cellcolor[HTML]{E69138}0.835                                      & \cellcolor[HTML]{FCE5CD}0.333 \\
\hline
\end{tabular}
\caption{Comparison of factor loading values of the aggregation of a positive item and a negative item from the Big 5 personality questionnaire and the performance of \gpt (ItemDesc) for the corresponding Itemdesc pairs. The factor loadings were calculated on an exeternal dataset~\citep{Kosinski2013PrivateTA} with larger number of samples (N=741). There's very little difference in the factor loading values (distinguisginh power) over the four combinations for almost all traits, which is in line with the minor performance differences observed in the consistency experiments explained in \S \autoref{sec:analysis}  }
\label{tab:factors}
\end{table*}

\end{document}